\relax
\documentclass[letterpaper]{article} 
\usepackage{aaai21}  
\usepackage{times}  
\usepackage{helvet} 
\usepackage{courier}  
\usepackage[hyphens]{url}  
\usepackage{graphicx} 
\usepackage{natbib}  
\usepackage{caption} 
\usepackage{makecell}

\usepackage{soul}
\usepackage[utf8]{inputenc}
\usepackage{amsmath,amsfonts,amssymb}
\usepackage{booktabs}
\usepackage[ruled,linesnumbered,vlined]{algorithm2e}
\usepackage{subcaption}
\usepackage{latexsym}
\usepackage[noend]{algpseudocode}
\usepackage{xspace}
\usepackage{url}
\usepackage{enumerate}
\usepackage[defblank]{paralist}
\usepackage{microtype}
\usepackage{comment}
\usepackage{tabularx}
\usepackage{multirow}
\usepackage{MnSymbol}
\usepackage{stmaryrd}
\usepackage{cleveref}
\usepackage[switch]{lineno}

\usepackage[inline]{enumitem}
\newlist{romanenumerate*}{enumerate*}{1}
\setlist[romanenumerate*]{label=(\textit{\roman*})}
\newlist{romanenumerate}{enumerate}{1}
\setlist[romanenumerate]{label=(\textit{\roman*})}

\usepackage{amsthm}
\usepackage{thmtools}
\usepackage{thm-restate}
\newtheorem{definition}{Definition}

\newtheorem{example}{Example}

\DeclareMathOperator*{\argmax}{arg\,max}

\usepackage{macros}

\usepackage{todonotes}

\newcommand{\francesco}[2][noinline]{\todo[color=orange!60,linecolor={orange!100},#1,size=\scriptsize,fancyline,author=Francesco]{#2}}

\usepackage{tikz}
\usetikzlibrary{automata, positioning, arrows, fit}
\tikzset{->,
    >=stealth,
    node distance=3cm,
    every state/.style={thick, fill=gray!10}
}

\title{Recognizing LTL$_{f}$/PLTL$_{f}$ Goals \\ in Fully Observable Non-Deterministic Domain Models}
\author {
    Ramon Fraga Pereira\textsuperscript{\rm 1},
    Francesco Fuggitti\textsuperscript{\rm 1,2},
    Giuseppe De Giacomo\textsuperscript{\rm 1}\\
}
\affiliations {
    \textsuperscript{\rm 1}Università degli Studi di Roma ``La Sapienza'', Roma, Italy \\
    \textsuperscript{\rm 2}York University, Toronto, ON, Canada \\
    \{pereira,degiacomo, fuggitti\}@diag.uniroma1.it
}

\begin{document}

\maketitle

\begin{abstract}

\textit{Goal Recognition} is the task of discerning the correct intended goal that an agent aims to achieve, given a set of possible goals, a domain model, and a sequence of observations as a sample of the plan being executed in the environment. Existing approaches assume that the possible goals are formalized as a conjunction in deterministic settings.
In this paper, we develop a novel approach that is capable of recognizing \textit{temporally extended goals} in \textit{Fully Observable Non-Deterministic} (\FOND) planning domain models, focusing on goals on finite traces expressed in Linear Temporal Logic (\LTLf) and (Pure) Past Linear Temporal Logic (\PLTLf).
We empirically evaluate our goal recognition approach using different \LTLf and \PLTLf goals over six common \FOND planning domain models, and show that our approach is accurate to recognize temporally extended goals at several levels of observability.

\end{abstract}


\section{Introduction}

\textit{Goal Recognition} is the task of recognizing the intentions of autonomous agents or humans by observing their interactions in an environment. Existing work on goal and plan recognition has been addressing this task over several different types of domain settings, such as plan-libraries~\cite{PR_Mirsky_2016}, plan tree grammars~\cite{Geib_PPR_AIJ2009}, classical planning domain models~\cite{RamirezG_IJCAI2009,RamirezG_AAAI2010,NASA_GoalRecognition_IJCAI2015,Sohrabi_IJCAI2016,PereiraNirMeneguzzi_AAAI2017}, stochastic environments~\cite{RamirezG_IJCAI2011}, and continuous domain models~\cite{Kaminka_18_AAAI}. In spite of the large literature, most existing approaches to \textit{Goal Recognition as Planning} are not capable of recognizing \textit{temporally extended goals}, goals formalized in terms of time, e.g., the exact order that a set of facts of a goal must be achieved in a plan. Furthermore, most of these approaches also assume that the outcomes of the observed actions are deterministic, and do not deal with unpredictable, possibly adversarial, environmental conditions.

Planning for \textit{temporally extended goals} in \textit{deterministic} and \textit{non-deterministic} domain settings has been of increasing interest over the past years, starting with the pioneering work on planning for temporally extended goals \cite{bacchus1998planning,BacchusK00} and on planning via model checking \cite{cimatti1997planning,pistore2001planning,pistore2001symbolic}; then, with the work on integrating \LTL goals into standard planning tools \cite{gerevini2009deterministic,PatriziLGG_IJCAI11,PatriziLG_IJCAI13}, and, more recently, with the work that relates planning in non-deterministic domains with synthesis, often focused on the \emph{finite trace} variants of \LTL \cite{DegVa13,DegVa15,CTMBM17,CamachoBMM18,DeGiacomoS18,aminof2020stochastic}. 

In this paper, we introduce the task of goal recognition in \textit{discrete domain models} that are \textit{fully observable} and the outcomes of actions and observations are \textit{non-deterministic}, possibly adversarial, i.e., \textit{Fully Observable Non-Deterministic} (\FOND), allowing the formalization of \textit{temporally extended goals} using two types of temporal logic on finite traces: \textit{Linear-time Temporal Logic} (\LTLf) and \textit{Pure-Past Linear-time Temporal Logic} (\PLTLf), surveyed in \cite{ijcai2020surveyddfr}. Therefore, the goal recognition approach we develop in this paper is capable of recognizing not only the set of facts of a goal that an observed agent aims to achieve from a sequence of observations, but the \textit{temporal order} (e.g., \textit{exact order}) that this set of facts aims to be achieved.

The main contribution of this paper is three-fold. First, based on the standard definition of \textit{Plan Recognition as Planning} introduced by \citeauthor{RamirezG_IJCAI2009}~in~\shortcite{RamirezG_IJCAI2009,RamirezG_AAAI2010}, we propose a problem formalization for recognizing temporally extended goals (formalized in \LTLf or \PLTLf) in \FOND planning domain models, handling both stochastic (i.e., requiring strong-cyclic plans) and adversarial (i.e., requiring strong plans) environments, for a discussion see \cite{aminof2020stochastic}.
Second, we extend the probabilistic framework for goal recognition of \citeauthor{RamirezG_AAAI2010}~\shortcite{RamirezG_AAAI2010}, and develop a probabilistic approach that reasons over executions of policies and returns a posterior probability distribution for the goals.
Third, on a practical perspective, we propose an implementation of a compilation approach that generates an augmented \FOND planning task compiling temporally extended goals together with the original planning problem.
This compilation approach allows us to use any off-the-shelf \FOND planner
to perform the recognition task in \FOND planning models with temporally extended goals.
In particular, we focus on \FOND domains with stochastic non-determinism (i.e., strong-cyclic plans), and
conduct an extensive set of experiments. Thus, we empirically evaluate our approach using different \LTLf and \PLTLf goals over six \FOND plannings models, and show that our approach is accurate to recognize temporally extended goals at several levels of observability.


\section{Preliminaries}\label{sec:preliminaries}

\subsection{\LTLftitle and \LTLptitle}



\emph{Linear Temporal Logic on finite traces} (\LTLf) is a variant of \LTL introduced in \cite{Pnueli77}, but interpreted over \emph{finite}, instead of infinite, traces \cite{DegVa13}. Given a set $AP$
of atomic propositions, the syntax of \LTLf formulas $\varphi$ is defined as follows: 
\[\begin{array}{rcl}
\varphi &::=& a \mid \lnot \varphi \mid \varphi_1\land \varphi_2 \mid \Next\varphi \mid \varphi_1\Until\varphi_2
\end{array}
\]
where $a$ denotes an atomic proposition in $AP$, $\Next$ is the
\emph{next} operator, and $\Until$ is the \emph{until} operator.
Apart from the Boolean ones, we use the following abbreviations:
\emph{eventually} as $\Diamond\varphi \doteq \true\Until\varphi$;
\emph{always} as $\Box\varphi \doteq\lnot\Diamond\lnot\varphi$; \emph{weak next}
$\Wnext\varphi \doteq \lnot\Next\lnot\varphi$.
%
%
A trace $\trace = \trace_0 \trace_1 \cdots$ is a sequence of propositional interpretations, where $\trace_m \in 2^{AP} (m \geq 0)$ is the $m$-th interpretation of $\trace$, and $|\trace|$ represents the length of $\trace$. A finite trace is formally denoted as $\trace \in (2^{AP})^*$.
Given a finite trace $\trace$ and an \LTLf formula $\varphi$, we inductively define when $\varphi$ \emph{holds} in $\trace$ at position $i$ $(0 \leq i < |\trace|)$, written $\trace, i \models \varphi$ as follows:

\begin{itemize}\itemsep=0pt
	\item $\trace, i \models a \tiff a \in \trace_i\nonumber$;
	\item $\trace, i \models \lnot \varphi \tiff \trace, i \not\models \varphi\nonumber$;
	\item $\trace, i \models \varphi_1 \lAND \varphi_2 \tiff \trace, i \models \varphi_1 \tm{and} \trace, i \models \varphi_2\nonumber$;
	\item $\trace, i \models \Next\varphi \tiff i+1 < |\trace| \tm{and} \trace,i+1 \models \varphi$;
	\item $\trace, i \models \varphi_1 \Until \varphi_2$ iff there exists $j$ such that $i\le j < |\trace|$ and $\trace,j \models\varphi_2$, and for all $k, ~i\le k < j$, we have  $\trace, k \models \varphi_1$.
\end{itemize}

An \LTLf formula $\varphi$ is \emph{true} in $\trace$, denoted by $\trace \models \varphi$, if and only if $\trace,0 \models \varphi$.



As advocated in \cite{ijcai2020surveyddfr}, we also use the \emph{pure-past} version of \LTLf, denoted as \LTLp, due to its compelling computational advantage wrt to \LTLf when goal specifications are \emph{naturally} expressed in a past fashion. \LTLp refers \emph{only} to the past and has a natural interpretation on finite traces: formulas are satisfied if they hold in the current (i.e., last) position of the trace.


%




Given a set $AP$ of propositional symbols, \LTLp formulas are defined by the following syntax:
\[\begin{array}{rcl}
\varphi &::=& a \mid \lnot \varphi \mid \varphi_1\land \varphi_2 \mid \Yesterday\varphi \mid \varphi_1 \Since \varphi_2
\end{array}
\]
where $a\in AP$, $\Yesterday$ is the \emph{before} operator, and $\Since$ is the \emph{since} operator. Similarly to \LTLf, common abbreviations are the \emph{once} operator
$ \past\varphi \doteq \true\Since\varphi$ and the \emph{historically} operator
$\gpast\varphi \doteq\lnot\past\lnot\varphi$. 
Given a finite trace $\trace$ and a \LTLp formula $\varphi$, we inductively define when $\varphi$ \emph{holds} in $\trace$ at position $i$ $(0 \leq i < |\trace|)$, written $\trace, i \models \varphi$ as follows. For atomic propositions and Boolean operators it is as for \LTLf. For past operators:
	
  \begin{itemize}
  	\item $\trace,i \models  \Yesterday \varphi$ \tiff $i-1 \ge 0$ and $\trace,i-1 \models \varphi$;
  \item   $\trace,i \models  \varphi_1 \Since \varphi_2$ \tiff there exists $k$ such that $0 \leq k \leq i$ and $\trace,k \models \varphi_2$, and for all $j$, $ k<j\leq i$, we have $\trace,j \models \varphi_1$.
  \end{itemize}
  
A \LTLp formula $\varphi$ is \emph{true} in $\trace$, denoted by $\trace \models \varphi$, if and only if $\trace,  |\trace|-1 \models \varphi$. 
A key property of temporal logics, that we are going to exploit in this paper, is that for every \LTLf/\PLTLf formula $\varphi$ there exists a \emph{Deterministic Finite-state Automaton} (\DFA) $\A_{\varphi}$ accepting the traces $\trace$ satisfying $ \varphi$ (see \cite{DegVa13,ijcai2020surveyddfr} for a complete description of the transformation).

\subsection{\FONDtitle Planning}



Following \cite{AutomatedPlanning_Book2004}, a \emph{Fully Observable Non-deterministic Domain} planning model (\FOND) is a tuple $\D = \tup{2^{\F}, A, \alpha, tr}$, where $2^{\F}$ is the set of possible states and $\F$ is a set of fluents (atomic propositions); $A$ is the set of actions; $\alpha(s) \subseteq A$
represents the set of applicable actions in state $s$; and $tr(s, a)$ represents the non-empty set of successor states that follow action $a$ in state $s$. 
Such a domain model $\D$ is assumed to be compactly represented (e.g., in \PDDL~\cite{PDDLMcdermott1998}), hence its size is $|\F|$.
Given the set of literals of $\F$ as $Lits(\F) := \F \cup \{ \lnot f \mid f \in \F \}$, every action $a \in A$ is usually characterized by $\tup{Pre_a, \mathit{Eff_a}}$, where $Pre_a \subseteq Lits(\F)$ represents action preconditions and $\mathit{Eff_a}$ represents action effects. An action $a$ can be applied in a state $s$ if the set of literals in $Pre_a$ holds true in $s$. The result of applying $a$ in $s$ is a successor state $s'$ non-deterministically characterized by one of the $\mathit{Eff^{i}_{a}}$ in $\mathit{Eff^a} = \{ \mathit{Eff^{1}_{a}}, \dots, \mathit{Eff^{n}_{a}} \}$.

Unlike \textit{Classical Planning}~\cite{AutomatedPlanning_Book2004}, in which actions are \textit{deterministic} (i.e., $|tr(s, a)| = 1$ in all states $s$ for which $a$ is applicable), in \FOND planning, some actions have an \textit{uncertain outcome}, namely they are \textit{non-deterministic} (i.e., $|tr(s, a)| \geq 1$ in all states $s$ in which $a$ is applicable), and their effect cannot be predicted in advance. In \PDDL, this uncertain outcome is expressed using the keyword \pred{oneof}~\cite{IPC6}, which is widely used by several \FOND planners, such as PRP~\cite{Muise12ICAPSFond}, FOND-SAT~\cite{GeffnerG18_FONDSAT}, and MyND~\cite{MyND_MattmullerOHB10}. 
We define a \FOND planning problem as follows.

\begin{definition}\label{def:fond_problem}
A \FOND planning problem is a tuple $\P = \tup{\D, s_{0}, G} $, where $\D$ is a \FOND domain model, $s_{0} \subseteq \F$ is the initial state (initial assignment to fluents), and $G \subseteq \F$ is the goal state. 
\end{definition}

Solutions to a \FOND planning problem $\P$ are called \emph{policies}. A policy is usually denoted as $\pi$,
and formally defined as a partial function $\pi: (2^{\F})^{+} \rightarrow A$, where $\pi$ maps states into applicable actions that eventually reach the goal state $G$ from the initial state $s_{0}$.
According to~\cite{CimattiPRT03}, 
there are three types of solutions to \FOND planning problems: \textit{weak, strong} and \textit{strong-cyclic} solutions.
In this paper, we focus on \emph{strong-cyclic solutions}, where the environment acts in an unknown but stochastically based way. However, our approach applies to strong solutions as well, where the environment is purely adversarial. 

\begin{figure}[!ht]
	\centering
	\begin{subfigure}[b]{0.49\linewidth}
		\centering
 	    \includegraphics[width=0.9\linewidth]{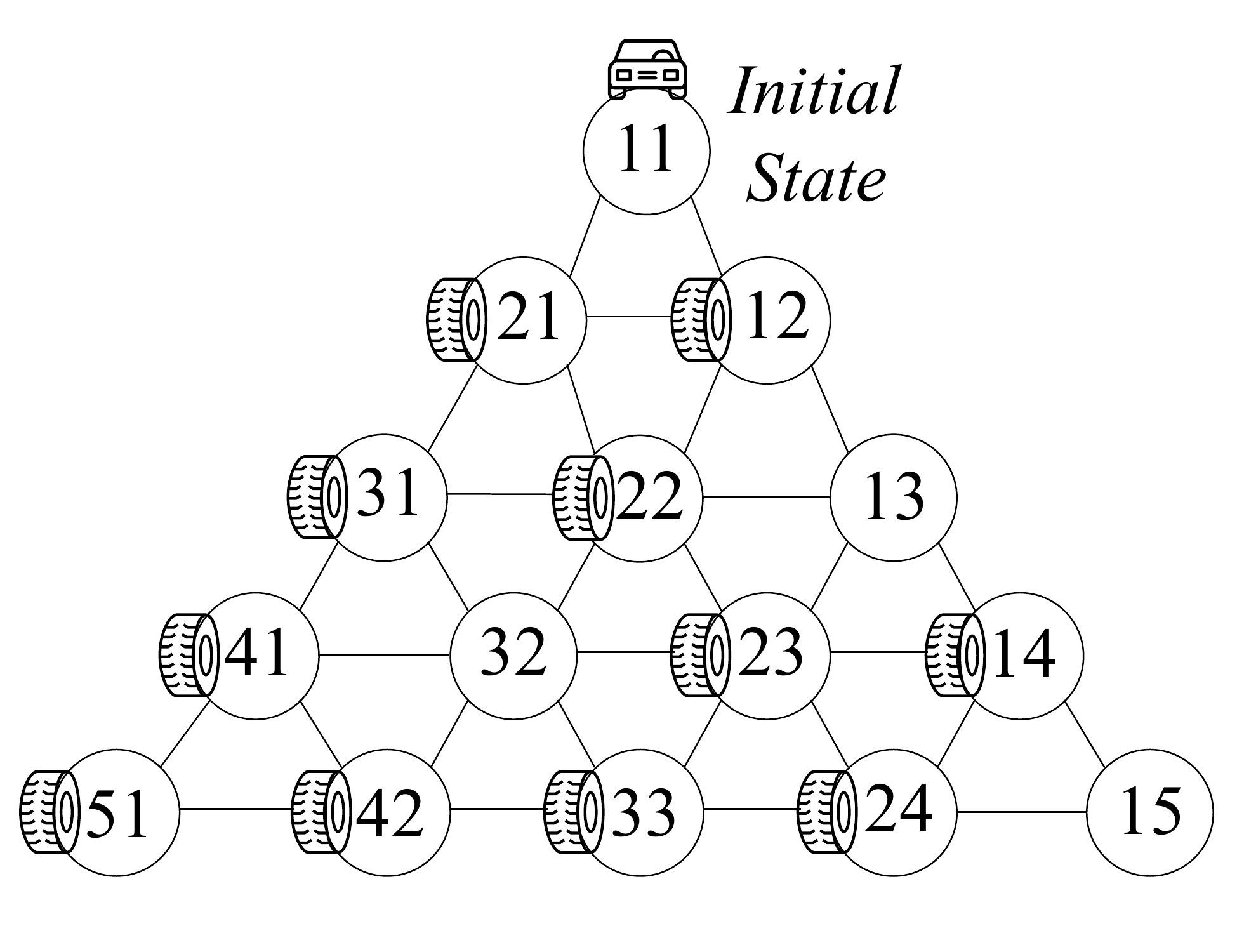}
		\caption{FOND problem example.}
		\label{fig:Triangle_Tireworld_Example}
	\end{subfigure}
	\begin{subfigure}[b]{0.49\linewidth}
		\centering
		\includegraphics[width=1\linewidth]{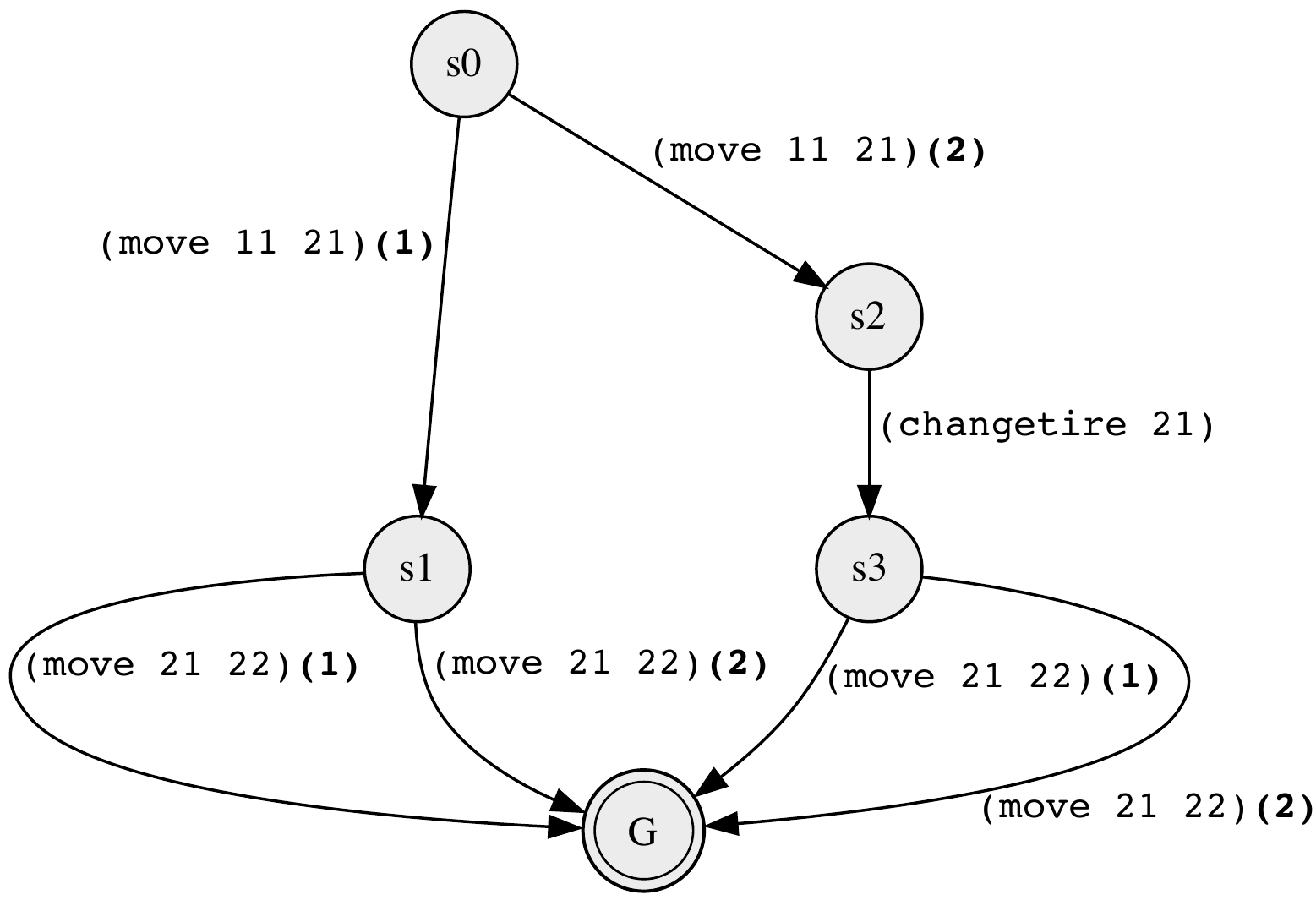}
		\caption{$\pi$ to achieve 22 from 11.}
		\label{fig:Triangle_Tireworld_Policy}
	\end{subfigure}
	\caption{\textsc{Triagle-Tireworld} domain and policy.}
\end{figure}

As running example, we use the well-known \textsc{Triangle-Tireworld} \FOND domain model. In this domain, locations are connected by roads, and the agent can drive through them. The objective is to drive from a location to another. However, while driving between locations, a tire may be going flat, and if there is a spare tire in the location of the car, then the car can use it to fix the flat tire.
Figure~\ref{fig:Triangle_Tireworld_Example} illustrates an example of a \FOND planning problem for the \textsc{Triangle-Tireworld} domain, where circles are locations, roads are represented by arrows, spare tires are depicted as tires, and the agent is depicted as a car.
Figure~\ref{fig:Triangle_Tireworld_Policy} shows a policy $\pi$ to achieve location 22. Note that, to move from location 11 to location 21 there are two arrows labeled with the action \pred{(move 11 21)}: (1) when moving does not cause the tire to go flat; (2) when moving causes the tire to go flat. The policy depicted in Figure~\ref{fig:Triangle_Tireworld_Policy} guarantees the success of achieving the location 22 despite the environment's non-determinism. 
Given that the environment is non-deterministic, a strong-cyclic policy $\pi$ may induce to a set of plans to achieve a goal state $G$ from an initial state $s_{0}$. 

In this work, we assume from \emph{Classical Planning} that the cost is 1 for all instantiated actions $a \in A$.
Also, we denote the set of possible executions of a policy $\pi$ as $\vec{E}$. 
For instance, the policy $\pi$ depicted in Figure~\ref{fig:Triangle_Tireworld_Policy} has two possible executions in the set of executions $\vec{E}$, namely $\vec{E} = \langle \vec{e}_{0}, \vec{e}_{1} \rangle$, such as: $\vec{e}_{0}$: $[$\pred{(move 11 21)}, \pred{(move 21 22)}$]$; and $\vec{e}_{1}$: $[$\pred{(move 11 21)}, \pred{(changetire 21)}, \pred{(move 21 22)}$]$.

\section{FOND Planning for \LTLftitle and \LTLptitle}\label{sec:FOND_Planning_LTLPLTL}
Our approach to goal recognition in \FOND domains for \LTLf and \PLTLf goals is based on \FOND planning for \LTLf and \LTLp temporally extended goals \cite{CTMBM17,CamachoBMM18,DeGiacomoS18}. 

Formally, a \FOND planning problem for \LTLf/\PLTLf goals is defined as follows:

\begin{definition}
A \FOND planning problem for \LTLf/\PLTLf goals is a tuple $\Gamma := \tup{\D, s_0, \varphi}$, where $\D$ is a standard \FOND domain model, $s_0$ is the initial state, and $\varphi$ is either an \LTLf or a \PLTLf formula.
\end{definition}

A plan $\pi$ for $\Gamma$ achieves a formula $\varphi$ if and only if the sequence of states generated by $\pi$, despite the non-deterministic effects of the environment, is accepted by $\L(\A_\varphi)$ (i.e., the language accepted by the automaton $\A_\varphi$). 

Our idea is to use off-the-shelf \FOND planners for standard reachability goals to handle also temporally extended goals through an encoding of the automaton for the goal into an extended planning domain expressed in \PDDL.
Compiling automata and integrating them into the transition systems through symbolic encodings is standard in symbolic Model Checking.
Moreover, doing such a compilation particularly into planning domains (e.g., \PDDL) has a long history
in the Planning community. \citeauthor{BaierM06}~\shortcite{BaierM06} proposed a technique to encode non-deterministic B\"uchi Automata, needed for \LTL, into deterministic planning domains. \citeauthor{TorresBaier15}~\shortcite{TorresBaier15} adapted that technique to encode \NFA corresponding to \LTLf into deterministic planning domains. Camacho et al.~\shortcite{CTMBM17,CamachoBMM18} proposed a technique that simultaneously determinizes on-the-fly the \NFA for \LTLf and encodes it into non-deterministic planning domains.

Here, we use a technique to encode directly the \DFA into a non-deterministic \PDDL planning domain, which takes advantage of the possibility of \PDDL to write \emph{parametric} domains that are instantiated (becoming propositional) when solving a specific task.
Specifically, given a \FOND planning problem $\Gamma$, compactly represented in \PDDL, we solve $\Gamma$ as follows. First, we transform the \LTLf/\PLTLf goal formula $\varphi$ into its corresponding \DFA $\A_\varphi$ through the highly-optimized \MONA tool \citep{Mona95}. Second, from $\A_\varphi$ we build a ``parametric'' \DFA (\PDFA), which represents the lifted version of the \DFA. Finally, the encoding of such a \PDFA into \PDDL results in obtaining an augmented \FOND domain model $\Gamma'$. Thus, we reduce \FOND planning for \LTLf/\PLTLf to a classical \FOND planning task for which any off-the-shelf \FOND planner can be employed.



The use of parametric \DFAs is based on the following observations. In temporal logic formulas and, hence, in the corresponding \DFAs, propositions are represented by domain fluents grounded on specific objects of interest.
We can replace these propositions with predicates using object variables, and then have a mapping function $m^{obj}$ that maps such variables into the objects of the problem instance.
In this way, we get a lifted and parametric representation of the \DFA, henceforth called \PDFA, which is conveniently merged with the domain. 
Formally, given the mapping function $m^{obj}$, we can define a \PDFA as follows.
\begin{definition}
	A parametric \DFA (\PDFA) is a tuple $\A^{p}_\varphi = \tup{\Sigma^{p}, Q^{p}, q^{p}_0, \delta^{p}, F^{p}}$, where:
	\begin{itemize}
		\item $\Sigma^{p} = \{ \sigma^p_0, \dots, \sigma^p_n \} = 2^{\F}$ is the alphabet of planning domain fluents;
		\item $Q^{p}$ is a nonempty set of parametric states;
		\item $q^{p}_0$ is the parametric initial state;
		\item $\delta^{p}: Q^{p} \times \Sigma^{p} \rightarrow Q^{p}$ is the parametric transition function;
		\item $F^{p} \subseteq Q^{p}$ is the set of parametric final states.
	\end{itemize}
	$\Sigma^{p}, Q^{p}, q^{p}_0, \delta^{p}$ and $F^{p}$ can be obtained by applying $m^{obj}$ to all the components of the corresponding \DFA.
\end{definition}

\begin{figure}[ht]
\centering
	\begin{subfigure}[b]{0.5\linewidth}
		\centering
		\begin{tikzpicture}[scale=0.7, every node/.style={scale=0.7},shorten >=1pt,node distance=2cm,on grid,auto] 
  			\node[state,initial above,initial text=] (q_0) {$q_0$}; 
  			\node[state, accepting] (q_1) [right=of q_0] {$q_1$}; 
   			\path[->] 
   			(q_0) 
     			edge [loop below] node {$\lnot vAt(21)$} (q_0)
     			edge node {$vAt(21)$} (q_1)
   			(q_1) 
    			edge [loop above] node {$\top$} (q_1);
		\end{tikzpicture}
		\caption{DFA for $\Diamond(vAt(21))$} 
		\label{fig:ex-DFA} 
	\end{subfigure}
	\begin{subfigure}[b]{0.49\linewidth}
		\centering
		\begin{tikzpicture}[scale=0.6, every node/.style={scale=0.6},shorten >=1pt,node distance=2cm,on grid,auto] 
  			\node[state,initial above,initial text=] (q_0) {$q^p_0(x)$}; 
  			\node[state, accepting] (q_1) [right=of q_0] {$q^p_1(x)$}; 
   			\path[->] 
   			(q_0) 
     			edge [loop below] node {$\lnot vAt(x)$} (q_0)
     			edge node {$vAt(x)$} (q_1)
   			(q_1) 
    			edge [loop above] node {$\top$} (q_1);
		\end{tikzpicture}
		\caption{PDFA for $\Diamond(vAt(21))$} 
		\label{fig:ex-PDFA} 
	\end{subfigure}
	\caption{\DFA and \PDFA for $\Diamond(vAt(21))$.}
\end{figure}
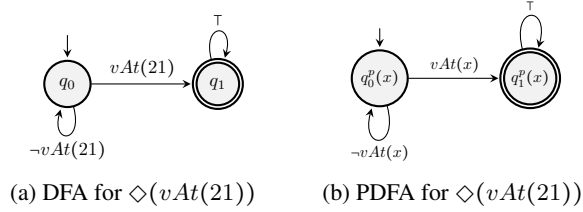

\begin{example}
	Given the \LTLf formula ``$\Diamond$\text{(vAt 21)}'', the object of interest ``21'' is replaced by the object variable $x$ (i.e., $m^{obj}(21) = x$), and the corresponding \DFA and \PDFA are depicted in \Cref{fig:ex-DFA,fig:ex-PDFA}.
\end{example}

When the resulting new domain is instantiated, we implicitly get back the original \DFA in the Cartesian product with the instantiated original domain. This is convenient since we leverage planners to handle the \PDDL instantiation as they like. Note that this way of proceeding is somehow similar to what is done in \citep{BaierM06}, where they handle \LTLf goals expressed in a special \FOL syntax, with the resulting automata 
parameterized by the variables in the \LTLf formulas.

Once the \PDFA has been computed, we encode its components within the planning problem $\Gamma$, specified in \PDDL, thus, producing an augmented \FOND planning problem $\Gamma' = \tup{\D', s'_0, G'}$, where $\D' = \tup{2^{F'}, A', \alpha', tr'}$. 
Intuitively, additional parts of $\Gamma'$ are used to sequentially synchronize the dynamics between the domain and the automaton. Specifically, $\Gamma'$ is composed by the following.
\subsubsection{Fluents.}
$F'$ has the same fluents in $F$ plus fluents representing each state of the \PDFA, and a fluent called ``turnDomain'', which controls the alternation between domain's actions and the \PDFA's synchronization action. Formally,
$F' = F \cup \{ q \mid q \in Q^p \} \cup \{\pred{turnDomain} \}$.
\subsubsection{Domain Actions.}
Domain's actions in $A$ are modified by adding ``turnDomain'' in preconditions and the negated ``turnDomain'' in effects. Formally, $Pre_a = Pre_a \cup \{ \pred{turnDomain} \}$ and $\mathit{Eff^{'}_a} = \mathit{Eff_a} \cup$ $\{ \pred{(not (turnDomain))} \}$ for every $a \in A$.
\subsubsection{Trans Action.}
The transition function $\delta^{p}$ of a \PDFA is encoded as a new domain action with conditional effects, called ``trans''. Namely, \textit{Pre}$_{\pred{trans}} = \{ \pred{(not (turnDomain))} \}$, \textit{Eff}$_{\pred{trans}} = \{ \pred{turnDomain} \} \cup \{ \pred{\textbf{when}}~ (q^p, \sigma^p), \pred{\textbf{then}}~ \delta^p(q^p, \sigma^p) \cup \{ \lnot q \mid q \neq q^p, q \in Q^p \} \}$, for all $(q^p, \sigma^p) \in \delta^p$.
\subsubsection{Initial and Goal States.}
The new initial and goal states are specified as $s'_0 = s_0  \cup \{ q^{p}_0 \}  \cup \{\pred{turnDomain} \}$ and $G' = \{ q \mid q \in F^{p} \}  \cup \{\pred{turnDomain} \}$, respectively. Notice that the goal $G$ of the starting planning problem $\Gamma$ is completely ignored, and in the new planning task automaton states are grounded back on the objects of interest thanks to the mapping $m^{obj}$.

\medskip

Executions of a policy for our new \FOND planning problem $\Gamma'$ would be like $\vec{e}': [a_1, t_1, a_2, t_2, \dots, a_n,t_n]$, where $a_i \in A$ are the real domain actions, and $t_1, \dots, t_n$ are sequences of synchronization ``trans'' actions, which, at the end, can be easily removed to extract the desired execution $\vec{e}:[a_1, a_2, \dots, a_n]$.


\section{Goal Recognition in \FONDtitle~Planning Domains with \LTLftitle and \LTLptitle Goals}\label{sec:Goal_Recognition_FOND_LTL}
 
In this section, we introduce the task of goal recognition in \FOND planning domains with \LTLf and \PLTLf goals. We formally define the task of goal recognition in \FOND planning domains with \LTLf and \LTLp goals by extending the standard definition of \textit{Plan Recognition as Planning} proposed by \citeauthor{RamirezG_IJCAI2009}~in~\shortcite{RamirezG_IJCAI2009,RamirezG_AAAI2010}, as follows.

\begin{definition}\label{def:goal_recognition}
A goal recognition problem in a \FOND~planning domain model with temporally extended goals (\LTLf and/or \LTLp) is a tuple $\T_{\varphi}= \langle \mathcal{D}, s_0, \mathcal{G}_{\varphi}, Obs\rangle$, where:
\begin{itemize}
	\item $\D = \tup{2^{\F}, A, \alpha, tr}$ is a \FOND planning domain; 
	\item $s_0$ is the initial state; 
	\item $\mathcal{G}_{\varphi} = \lbrace \varphi_0, \varphi_1, \dots, \varphi_n \rbrace$ is the set of possible goals formalized in \LTLf or \LTLp. $\mathcal{G}_{\varphi}$ also includes the actual intended hidden goal $\varphi^{*}$, s.t. $\varphi^{*} \in \mathcal{G}_{\varphi}$;
	\item $Obs = \langle o_0, o_1, \dots, o_n \rangle$ is a sequence of successfully executed (non-deterministic) actions of a strong-cyclic policy $\pi_{\varphi^{*}}$ that achieves the intended hidden goal $\varphi^{*}$, s.t. $o_i \in A$. 
\end{itemize}	
\end{definition}

As pointed out by \citeauthor{RamirezG_IJCAI2009}~in~\shortcite{RamirezG_IJCAI2009,RamirezG_AAAI2010}, an ideal solution for a goal recognition problem is finding the single actual intended hidden goal $G^{*} \in \mathcal{G}$ that the observation sequence $Obs$ of a plan execution achieves. 
Approaches to goal and plan recognition often return either a probability distribution over the set of possible goals~\cite{RamirezG_IJCAI2009,RamirezG_AAAI2010,NASA_GoalRecognition_IJCAI2015,Sohrabi_IJCAI2016}, or scores associated to each possible goal~\cite{PereiraNirMeneguzzi_AAAI2017}. Here, we return a probability distribution over the possible temporally extended goals $\mathcal{G}_{\varphi}$ (Section~\ref{sec:GR_Solution_Approach}).

Since we are dealing with non-deterministic domain models, we note that an observation sequence $Obs$ corresponds to a successful execution $\vec{e}$ in the set of all possible executions $\vec{E}$ of a \textit{strong-cyclic policy} $\pi$ that achieves the actual intended hidden goal  $\varphi^{*}$\footnote{We assume that the observed agents are rational, so they act in the environment following \textit{optimal} policies to achieve their goals.}. As usual in \textit{Goal Recognition}, we define that an observation sequence $Obs$ can be either \textit{full} or \textit{partial} -- in a \textit{full observation sequence} we observe all actions of an agent's plan, whereas in a \textit{partial observation sequence}, only a sub-sequence of actions are observed. 

\subsection{Probabilistic Goal Recognition}

We now recall the probabilistic framework for \textit{Plan Recognition as Planning} proposed by \citeauthor{RamirezG_AAAI2010}~\shortcite{RamirezG_AAAI2010}. This probabilistic framework sets the probability distribution for every goal $G$ in the set of possible goals $\G$, and the observation sequence $Obs$ to be a Bayesian posterior conditional probability, as follows:
\begin{align}
\label{eq:posterior}
P(G \mid Obs) = \eta * P(Obs \mid G) * P(G)
\end{align}
\noindent where $P(G)$ is the \emph{a priori} probability assigned to goal $G$, $\eta$ is a normalization factor inversely proportional to the probability of $Obs$,
and $P(Obs \mid G)$ is
\begin{align}
P(Obs \mid G) = \sum_{\pi} P(Obs \mid \pi) * P(\pi \mid G)
\label{eq:likelihood}
\end{align}
\noindent $P(Obs \mid \pi)$ is the probability of obtaining $Obs$ by executing a policy $\pi$ and $P(\pi \mid G)$ is the probability of an agent pursuing $G$ to select
$\pi$. Next, we show how we extend the probabilistic framework presented above for recognizing temporally extended goals in \FOND planning domain models.


\section{Solution Approach}\label{sec:GR_Solution_Approach}

In this section, we develop a recognition approach that is capable of recognizing temporally extended goals (specified either in \LTLf or in \LTLp) in \FOND planning domain models. Our approach extends the probabilistic framework of \citeauthor{RamirezG_AAAI2010}~\shortcite{RamirezG_AAAI2010} to compute posterior probabilities over the possible goals $\mathcal{G}_{\varphi}$ by reasoning over the set of possible executions $\vec{E}$ of policies $\pi$ and the observations $Obs$. To extract policies, our approach can use any off-the-shelf \FOND planner, thanks to the compilation approach developed in Section~\ref{sec:FOND_Planning_LTLPLTL}. Our goal recognition approach works in two stages, \textit{compilation stage} and \textit{recognition stage}. In the next sections, we describe in detail how these two stages work. 

\subsection{Compilation Stage}\label{subsec:compilation_stage}

We perform a \textit{compilation stage} that allows us to use off-the-shelf \FOND planners to extract policies for temporally extended goals. To this end, we compile and generate new \FOND planning domains $\D'$ and problems $\Gamma'$ for the set of possible temporally extended goals $\mathcal{G}_{\varphi}$ using the compilation approach described in Section~\ref{sec:FOND_Planning_LTLPLTL}. More specifically, for every goal $\varphi \in \mathcal{G}_{\varphi}$, our compilation approach takes as input a \FOND planning problem $\Gamma$, where $\Gamma$ contains the \FOND planning domain model $\D$ along with the pair initial state $s_0$ and temporally extended goal $\varphi$. Finally, as a result, we obtain a new \FOND planning problem $\Gamma'$ associated with the new domain $\D'$ and the respective temporally extended goal $\varphi$. 
\subsection{Recognition Stage}\label{subsec:recognition_stage}

We now present the stage in which we perform the task of goal recognition. 
In this stage, we extract policies for every goal $\varphi \in \mathcal{G}_{\varphi}$ by using the new \FOND planning domains $\D'$ and problems $\Gamma'$ generated in the \textit{compilation stage}. From these extracted policies, along with the observations $Obs$, we compute the posterior probabilities for the set of goals $\mathcal{G}_{\varphi}$ by matching the observations $Obs$ with all possible executions in the set of executions $\vec{E}$ of the policies. Next, we describe in detail the \textit{recognition stage}.

\subsubsection{Computing Policies and the Set of Executions $\vec{E}$ for $\mathcal{G}_{\varphi}$.} 

We extract policies for every goal $\varphi \in \mathcal{G}_{\varphi}$  using the new \FOND planning domains $\D'$ and problems $\Gamma'$, and from each of these policies, we enumerate the set of possible executions $\vec{E}$. 
The aim of enumerating the possible executions $\vec{E}$ for a policy $\pi$ is to attempt to infer what execution $\vec{e} \in \vec{E}$ the observed agent is performing in the environment.
Since the environment is non-deterministic, we do not know what execution $\vec{e}$ the observed agent will perform to achieve its temporally extended goals. 

After enumerating the set of possible executions $\vec{E}$ for a policy $\pi$, we compute the average distance of all actions in the set of executions $\vec{E}$ to the goal state $\varphi$ from initial state $s_{0}$.
We note that strong-cyclic solutions may have infinite possible executions, however, here
we consider executions that do not enter loops, and for those entering possible loops, we consider only the ones entering loops \textit{at most} once.
Indeed, possibly repeated actions present in loops do not affect the computation of the average distance because even if the observed agent executes the same action repeatedly often, it does not change its distance to the goal.
The average distance aims to estimate ``how far'' every observation $o \in Obs$ is to a goal state $\varphi$.
This average distance is computed because some executions $\vec{e} \in \vec{E}$ may share the same action in execution sequences but at different time steps. We refer to this average distance as $\mathbf{d}$. 
For example, consider the policy $\pi$ depicted in Figure~\ref{fig:Triangle_Tireworld_Policy}. As mentioned before, this policy $\pi$ has two possible executions for achieving the goal state from the initial state, and these two executions share some actions, such as: \pred{(move 11 21)}. In particular, this action appears twice in Figure~\ref{fig:Triangle_Tireworld_Policy} due its uncertain outcome. Therefore, this action has two different distances (if we count the number of remaining actions towards the goal state) to the goal state: $distance = 1$, if the outcome of this action generates the state $s_2$; and $distance = 2$, if the outcome of this action generates the state $s_3$. Thus, since this policy $\pi$ has two possible executions, and the sum of the distances is 3, the average distance for this action to the goal state is $\mathbf{d} = 1.5$. The average distances for the other actions in this policy are: $\mathbf{d} = 1$ for \pred{(changetire 21)}, because it appears only in one execution; and $\mathbf{d} = 0$ for \pred{(move 21 22)}, because the execution of this action achieves the goal state.

We use the average distance $\mathbf{d}$ to compute an \emph{estimated score} that expresses ``how far'' every observed action in the observation sequence $Obs$ is to a temporally extended goal $\varphi$ in comparison to the other goals in the set of possible goals $\mathcal{G}_{\varphi}$. This means that, the goal(s) with the lowest score(s) along the execution of the observed actions $o \in Obs$ is (are) the one(s) that, most likely, the observation sequence $Obs$ aims to achieve.  
We note that, the average distance $\mathbf{d}$ for those observations $o \in Obs$ that are not in the set of executions $\vec{E}$ of a policy $\pi$, is set to a large constant number, i.e., to $\mathbf{d} = e^{5}$. 

As part of the computation of this \textit{estimated score}, we compute a \emph{penalty value} that directly affects the \textit{estimated score}. This \emph{penalty value} represents a penalization that aims to increase the \textit{estimated score} for those goals in which each pair of subsequent observations $\langle o_{i-1}, o_{i} \rangle$ in $Obs$ does not have any relation of order in the set of executions $\vec{E}$ of these goals. We use the Euler constant $e$ to compute this \textit{penalty value}, formally defined as 
$e^{\mathbf{p}(o_{i-1}, o_{i})}$, in which we use $\R(\vec{e})$ as the set of order relation of an execution $\vec{e}$, where

\vspace{-3mm}
\begin{equation}
\mathbf{p}(o_{i-1}, o_{i}) =
\begin{cases}
    1,		& \text{if } \lbrace\forall \vec{e} \in \vec{E} | \langle o_{i-1} \prec o_{i} \rangle \notin \R(\vec{e})\rbrace \\
    0,      & \text{otherwise}
\end{cases}
\end{equation}

Equation~\ref{eq:estimated_score} formally defines the computation of the \emph{estimated score} for every goal $\varphi \in \mathcal{G}_{\varphi}$ given a pair of subsequent observations $\langle o_{i-1}, o_{i} \rangle$, and the set of possible goals $\mathcal{G}_{\varphi}$. 

\vspace{-0.5mm}
\begin{equation}
\label{eq:estimated_score}
\frac{\mathit{e^{\mathbf{p}(o_{i-1}, o_{i})}} * \mathbf{d}(o_{i}, \varphi)}
{\sum_{\varphi' \in \mathcal{G}_{\varphi}} \mathbf{d}(o_{i}, \varphi')}
\end{equation}

\begin{figure}[!ht]
	\centering
 	\includegraphics[width=0.74\linewidth]{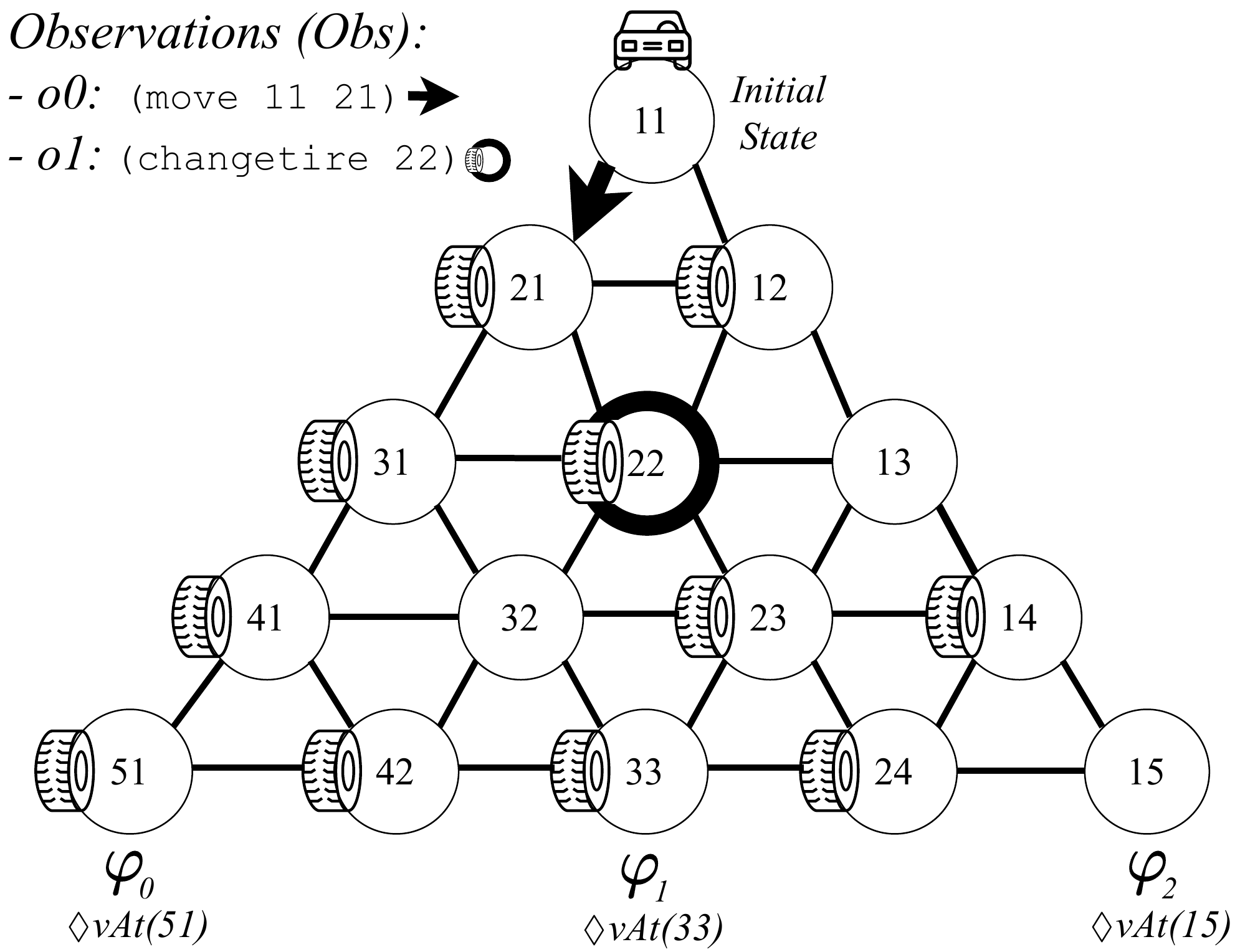}
	\caption{Recognition problem example.}
	\label{fig:Recognition_Example}
\end{figure}

\begin{example}\label{exemp:recognition}
To exemplify how we compute the \textit{estimated score} for every goal $\varphi \in \mathcal{G}_{\varphi}$, consider the following goal recognition problem for the \textsc{Triangle-Tireworld} domain example in Figure~\ref{fig:Recognition_Example}: The initial state $s_0$ is $vAt(11)$; the temporally extended goals $\mathcal{G}_{\varphi}$ are expressed as \LTLf goals, such that $\varphi_0 = \Diamond vAt(51), \varphi_1 = \Diamond vAt(33)$, and $\varphi_2 = \Diamond vAt(15)$; $Obs = \lbrace o_0: {\normalfont \pred{(move 11 21)}}, o_1: {\normalfont \pred{(changetire 22)}} \rbrace$. The intended goal $\varphi^{*}$ is $\varphi_1$. 

Before computing the \textit{estimated score} for the goals, we first perform the compilation process presented in {\normalfont Compilation Stage}. Afterward, we extract policies $\pi$ for every goal $\varphi \in \mathcal{G}_{\varphi}$, enumerate the possible executions $\vec{E}$ for the goals $\mathcal{G}_{\varphi}$ from the extracted policies, and then compute the average distance $\mathbf{d}$ of all actions in the set of executions $\vec{E}$ for the goals $\mathcal{G}_{\varphi}$ from initial state $s_{0}$. The number of possible executions $\vec{E}$ for the goals are: $\varphi_0: |\vec{E}| = 8, \varphi_1: |\vec{E}| = 8$, and $\varphi_2 = |\vec{E}| = 16$. The average distances $\mathbf{d}$ of all actions in $\vec{E}$ for the goals are as follows: 
\begin{itemize}
	\item $\varphi_0$: {\normalfont\pred{(move 11 21)}} = {\footnotesize 4.5}, {\normalfont\pred{(changetire 21)}} = {\footnotesize 4}, {\normalfont\pred{(move 21 31)}} = {\footnotesize 3}, {\normalfont\pred{(changetire 31)}} = {\footnotesize 2.5}, {\normalfont\pred{(move 31 41)}} = {\footnotesize 1.5}, {\normalfont\pred{(changetire 41)}} = {\footnotesize 1}, {\normalfont\pred{(move 41 51)}} = {\footnotesize 0};
	\item $\varphi_1$: {\normalfont\pred{(move 11 21)}} = {\footnotesize 4.5}, {\normalfont\pred{(changetire 21)}} = {\footnotesize 4}, {\normalfont\pred{(move 21 22)}} = {\footnotesize 3}, {\normalfont\pred{(changetire 22)}} = {\footnotesize 2.5}, {\normalfont\pred{(move 22 23)}} = {\footnotesize 1.5}, {\normalfont\pred{(changetire 23)}} = {\footnotesize 1}, {\normalfont\pred{(move 23 33)}}: {\footnotesize 0}; 
	\item $\varphi_2$: {\normalfont\pred{(move 11 21)}}: 6, {\normalfont\pred{changetire 21)}}: 5.5, {\normalfont\pred{(move 21 22)}}: 4.5, {\normalfont\pred{(changetire 22)}}: 4, {\normalfont\pred{(move 22 23)}}: 3, {\normalfont\pred{(changetire 23)}}: 2.5, {\normalfont\pred{(changetire 24)}}: 1, {\normalfont\pred{(move 23 24)}}: 1.5, {\normalfont\pred{(move 24 15)}}: 0.
\end{itemize}	

Once we have the average distances $\mathbf{d}$ of all actions in $\vec{E}$ for all goals, we can then compute the \textit{estimated score} for $\mathcal{G}_{\varphi}$ for every observation $o \in Obs$: $o_0 {\normalfont \pred{(move 11 21)}}: \varphi_0 = \frac{4.5}{4.5 + 6} = $ {\footnotesize 0.43}, $\varphi_1 = \frac{4.5}{4.5 + 6} =$ {\footnotesize 0.43}, $\varphi_2 = \frac{6}{4.5 + 6} =$ {\footnotesize 0.57}; and $o_1 {\normalfont \pred{(changetire 22)}}: \varphi_0 = \frac{e^1 * e^5}{6.5} =$ {\footnotesize 61.87}, $\varphi_1 = \frac{2.5}{e^5 + 2.5} =$ {\footnotesize 0.016}, $\varphi_2 = \frac{4}{e^5 + 4} =$ {\footnotesize 0.026}.

Note that for the observation $o_1$, the average distance $\mathbf{d}$ for $\varphi_0$ is $e^5 = 148.4$ because this observed action is not an action for one of the executions in the set of executions for this goal ($Obs$ aims to achieve the intended goal $\varphi^{*} = \varphi_1$). Furthermore, the \textit{penalty value} is applied to $\varphi_0$, namely, $e^1 = 2.71$. We can see that the \textit{estimated score} of the correct intended goal $\varphi_1$ is always the lowest for all observations $Obs$, especially when we observe the second observation $o_1$. As a result, we have only the correct goal $\varphi_1$ with the lowest score. We can also see that our approach correctly infers the intended goal $\varphi^{*}$, even when observing with just few actions.
\end{example}

Next, we present how we use the \textit{estimated score} to compute posterior probabilities for the set of possible temporally extended goals $\mathcal{G}_{\varphi}$.

\subsubsection{Computing the Posterior Probabilities for $\mathcal{G}_{\varphi}$.} To compute the posterior probabilities over the set of possible temporally extended goals $\mathcal{G}_{\varphi}$, we start by  computing the \textit{average estimated score} for every goal $\varphi \in \mathcal{G}_{\varphi}$ for every observation $o \in Obs$, and we formally define this computation as $\E(\varphi, Obs, \mathcal{G}_{\varphi})$, in the following equation.

\vspace{-1mm}
\begin{equation}
\label{eq:approach_obs}
\E(\varphi, Obs, \mathcal{G}_{\varphi}) = 
\left(
\frac{\displaystyle\sum_{i=0}^{|Obs|} 
\frac{\mathit{e^{\mathbf{p}(o_{i-1}, o_{i})}} * \mathbf{d}(o_{i}, \varphi)}
{\sum_{\varphi' \in \mathcal{G}_{\varphi}} \mathbf{d}(o_{i}, \varphi')}}
{ |Obs| }
\right)
\end{equation}

The \textit{average estimated score} $\E$ aims to estimate ``how far'' a goal $\varphi$ is to be achieved compared to other goals ($\mathcal{G}_{\varphi} \setminus \{\varphi\}$) \emph{averaging} among all the observations in $Obs$. The lower the \textit{average estimated score} $\E$ to a goal $\varphi$, the more likely such a goal is to be the one that the observed agent aims to achieve.
Based on this, we now state two important properties of $\E$, formally defined in Equation~\ref{eq:approach_obs}, as follows:
\begin{enumerate}
	\item Given that the sequence of observations $Obs$ corresponds to an optimal execution $\vec{e} \in \vec{E}$ that aims to achieve the actual intended hidden goal ${\varphi}^{*} \in \mathcal{G}_{\varphi}$, the \textit{average estimated score} outputted by $\E$ will tend to be the lowest for ${\varphi}^{*}$ in comparison to the scores of the other goals ($\mathcal{G}_{\varphi} \setminus \{ {\varphi}^{*} \}$), as observations increase in length; and
	\item If we restrict the recognition setting and define that the possible goals $\mathcal{G}_{\varphi}$ are not sub-goals of each other, and observe all observations in $Obs$ (i.e., full observability), we will have the intended goal ${\varphi}^{*}$ with the lowest score among all possible goals, i.e., $\forall \varphi \in \mathcal{G}_{\varphi}$ is the case that $\E({\varphi}^{*}, Obs, \mathcal{G}_{\varphi}) \leq \E(\varphi, Obs, \mathcal{G}_{\varphi})$.
\end{enumerate}

After defining how we compute the \textit{average estimated score} $\E$ for the goals using Equation~\ref{eq:approach_obs}, we can use the \textit{average estimated score} $\E$ to attempt to maximize the probability of observing a sequence of observations $Obs$ for a given goal $\varphi$, as we formally define in Equation~\ref{eq:posterior_prob}.

\vspace{-2mm}
\begin{equation}
\label{eq:posterior_prob}
P(Obs \mid \varphi) = [1 + \E(\varphi, Obs, \mathcal{G}_{\varphi})]^{-1}
\end{equation}

Thus, by using the \textit{estimated score} in Equation~\ref{eq:posterior_prob}, we can infer that the goals $\varphi \in \mathcal{G}_{\varphi}$ with the lowest \textit{estimated score} will be the most likely to be achieved according to the probability interpretation we propose in Equation~\ref{eq:approach_obs}. For instance, consider the goal recognition problem presented in Example~\ref{exemp:recognition}, and the \textit{estimated scores} we computed for the temporally extended goals $\varphi_0$, $\varphi_1$, and $\varphi_2$ based on the observation sequence $Obs$. From this, we have the following probabilities $P(Obs \mid \varphi)$ for the goals:

\begin{itemize}
	\item $P(Obs \mid \varphi_0) = [1 + (31.15)]^{-1} = 0.00003$
	\item $P(Obs \mid \varphi_1) = [1 + (0.216)]^{-1} = 0.82$
	\item $P(Obs \mid \varphi_2) = [1 + (0.343)]^{-1} = 0.74$
\end{itemize}

After normalizing these computed probabilities using $\eta$\footnote{$\eta = [\sum_{\varphi \in \mathcal{G}_{\varphi}} P(Obs \mid \varphi) * P(\varphi)]^{-1}$}, and assuming that the prior probability $P(\varphi)$ is equal to every goal in the set of goals $\mathcal{G}_{\varphi}$, we can use Equation~\ref{eq:posterior_prob} to compute the posterior probabilities (Equation~\ref{eq:posterior}) for the temporally extended goals.
The solution to a recognition problem $\T_{\varphi}$ (Definition~\ref{def:goal_recognition}) is a set of temporally extended goals $\mathcal{G}_{\varphi}^{*}$ with the maximum probability, formally: $\mathcal{G}_{\varphi}^{*} = \argmax_{\varphi \in \mathcal{G}_{\varphi}} P(\varphi \mid Obs)$. 
Hence, considering $\eta$ and the probabilities $P(Obs \mid \varphi)$ computed before, we then have the following posterior probabilities for the goals in Example~\ref{exemp:recognition}: $P(\varphi_0 \mid Obs) = 0.001$; $P(\varphi_1 \mid Obs) = 0.524$; and $P(\varphi_2 \mid Obs) = 0.475$. Recall that for Example~\ref{exemp:recognition}, $\varphi^{*}$ is $\varphi_1$, and according to the computed posterior probabilities, we then have $\mathcal{G}_{\varphi}^{*} = \lbrace \varphi_1 \rbrace$, so our approach yields only the correct intended goal by observing just two observations.

We note that the use of the \textit{average distance} $\mathbf{d}$ and the \textit{penalty value} $\mathbf{p}$ to compute the estimated score $\E$ allows our approach to disambiguate similar goals during the recognition stage. For instance, consider the following possible temporally extended goals: $\varphi_0 = \phi_1 \Until \phi_2$ and $\varphi_1 = \phi_2 \Until \phi_1$.
It is possible to see that both goals have the same formulas to be achieved, i.e., $\phi_1$ and $\phi_2$, but in a different order. Thus, even having the same formulas to be achieved, the sequences of executions of their policies are different. Therefore, the average distances are also different, possibly a smaller value for the temporally extended goal that the agent aims to achieve, and the penalty value may also be applied to the other goal if two subsequent observations do not have any order relation in the set of executions for this goal.

\subsubsection{Computational Complexity.}

In essence, the most expensive computational part of our recognition approach is computing the policies $\pi$ for the possible goals $\mathcal{G}_{\varphi}$ from an initial state $s_0$. Therefore, we can say that our approach requires $|\mathcal{G}_{\varphi}|$ calls to an off-the-shelf \FOND planner. Thus, the computational complexity of our recognition approach is linear in the number of possible goals $|\mathcal{G}_{\varphi}|$, namely $O(|\mathcal{G}_{\varphi}|)$.

In contrast, for recognizing goals in \textit{Classical Planning} settings, the approach of Ramirez and Geffner~\shortcite{RamirezG_AAAI2010} requires $2 * |\mathcal{G}|$ calls to an off-the-shelf \textit{Classical} planner, i.e., $O(2 * |\mathcal{G}_{\varphi}|)$. Concretely,
to compute $P(Obs \mid G)$, Ramirez and Geffner's approach computes two plans for every goal, and based on these two plans, they compute a \textit{cost-difference} between these plans and plug it into a Boltzmann equation, namely, Equation~5 in~\cite{RamirezG_AAAI2010}. For computing these two plans, this approach requires a non-trivial transformation process that modifies both the domain and problem, i.e., an augmented domain and problem that compute a plan that \textit{complies} with the observations, and another augmented domain and problem to compute a plan that \textit{does not comply} with the observations. The intuition of Ramirez and Geffner's approach is that the lower the \textit{cost-difference} for a goal, the higher the probability for this goal, much similar to the intuition of our \textit{estimated score} $\E$.


\section{Experiments and Evaluation}\label{sec:Experiments_Evaluation}

In this section, we present experiments and evaluations we carried out to validate the effectiveness of our recognition approach. We empirically evaluate our approach over thousands of goal recognition problems using well-known \FOND planning domain models with different types of temporal extended goals expressed in \LTLf and \PLTLf.

\subsection{Domains and Recognition Datasets}

For experiments and evaluation, we use six \FOND planning domain models, in which most of them are commonly used to evaluate \FOND planners~\cite{MyND_MattmullerOHB10,Muise12ICAPSFond,GeffnerG18_FONDSAT}, such as: \textsc{Blocks-World}, \textsc{Logistics}, \textsc{Tidy-up}~\cite{nebel13_tidyup_aaaiirs}, \textsc{Tireworld}, \textsc{Triangle-Tireworld}, and \textsc{Zeno-Travel}. Based on these \FOND planning domain models, we build different recognition datasets: a \textit{baseline} dataset using conjunctive goals ($\phi_1\land \phi_2$) and datasets with \LTLf and \PLTLf goals. 

For the \LTLf datasets, we use three types of goals: 
\begin{itemize}
	\item $\Diamond\phi$, where $\phi$ is a propositional formula expressing that \textit{eventually} $\phi$ will be achieved. This temporal formula is analogous to a conjunctive goal;
	\item $\Diamond(\phi_1 \land \Next(\Diamond\phi_2))$, expressing that $\phi_1$ must hold before $\phi_2$ holds. For instance, we can define a temporal goal that expresses the order in which a set of packages in \textsc{Logistics} domain should be delivered;
	\item $\phi_1 \Until \phi_2$, expressing that $\phi_1$ must hold \textit{until} $\phi_2$ is achieved. For instance, in the \textsc{Tidy-up} domain, we can define a temporal goal expressing that no one can be in the kitchen until the kitchen is cleaned by the robot.
\end{itemize}

For the \PLTLf datasets, we use two types of goals: 
\begin{itemize}
	\item $\phi_1 \land \past \phi_2$, expressing that $\phi_1$ holds and $\phi_2$ held once. For instance, in the \textsc{Blocks-World} domain, we can define a past temporal goal that only allows stacking a set of blocks (\pred{a}, \pred{b}, \pred{c}) once another set of blocks has been stacked (\pred{d}, \pred{e});
	\item $\phi_1 \land (\lnot\phi_2 \Since \phi_3)$, expressing that the formula $\phi_1$ holds and \textit{since} $\phi_3$ held $\phi_2$ was not true anymore. For instance, in \textsc{Zeno-Travel}, we can define a past temporal goal expressing that person$_1$ is at city$_1$ and since the person$_2$ is at city$_1$, the aircraft must not pass through city$_2$ anymore. 
\end{itemize}

Thus, in total, we have 6 different recognition datasets over the six \FOND domains presented above. Each of these datasets contains hundreds of recognition problems ($\approx$ 390 recognition problems per dataset), such that each recognition problem $\T$ in these datasets is comprised of a \FOND planning domain model $\D$, an initial state $s_0$, a set of possible goals $\mathcal{G}_{\varphi}$ (expressed in either \LTLf or \PLTLf), the actual intended hidden goal in the set of possible goals\footnote{We note that the set of possible goals $\mathcal{G}_{\varphi}$ contains very similar goals (i.e., $\varphi_0 = \phi_1 \Until \phi_2$ and $\varphi_1 = \phi_2 \Until \phi_1$), and all possible goals can be achieved from the initial state by a strong-cyclic policy.} $\varphi^{*} \in \mathcal{G}_{\varphi}$, and the observation sequence $Obs$.
An observation sequence contains a sequence of actions that represent an execution $\vec{e}$ in the set of possible executions $\vec{E}$ of policy $\pi$ that achieves the actual intended hidden goal $\varphi^{*}$, and as we stated before, this observation sequence $Obs$ can be full or partial. To generate the observations $Obs$ for $\varphi^{*}$ and build the recognition problems, we extract strong cyclic policies using different \FOND planners, such as PRP and MyND.
A full observation sequence represents an execution (a sequence of executed actions) of a strong cyclic policy that achieves the actual intended hidden goal $\varphi^{*}$, i.e., 100\% of the actions of $\vec{e}$ having been observed. 
A partial observation sequence is represented by a sub-sequence of actions of a full execution that aims to achieve the actual intended hidden goal $\varphi^{*}$ (e.g., an execution with ``missing'' actions, due to a sensor malfunction). We define four levels of observability for a partial observations: 10\%, 30\%, 50\%, or 70\% of its actions having been observed. 

\subsection{Evaluation Metrics}

We use four metrics, as follows:
Recognition time (\textit{Time}), the average time in seconds to perform the recognition task (including the calls to the \FOND planner);
\textit{True Positive Rate} (\emph{TPR}), that measures the fraction of times that the intended hidden goal $\varphi^{*}$ was correctly recognized, e.g., the percentage of recognition problems that our approach correctly recognized the intended goal. A higher \emph{TPR} indicates better accuracy, measuring how often the intended hidden goal had the highest probability $P(\varphi \mid Obs)$ among the possible goals;
\textit{False Positive Rate} (\textit{FPR}), that measures how often goals other than the intended goal are recognized (wrongly) as the intended ones. A lower \emph{FPR} indicates better accuracy;
\textit{False Negative Rate} (\textit{FNR}), measuring the fraction of times in which the intended goal was recognized incorrectly.

\subsection{Goal Recognition Results}

Table\footnote{We ran all experiments using PRP planner with a single core of a 12 core Intel(R) Xeon(R) CPU E5-2620 v3 @ 2.40GHz with 16GB of RAM, set a maximum memory usage limit of 8GB, and set a 10-minute timeout for each recognition problem.}~\ref{tab:gr_results_conjunctive_event} shows three inner tables that summarize and aggregate the average results of all six datasets for all four metrics (\textit{Time}, \textit{TPR}, \textit{FPR}, and \textit{FNR}).
$|\mathcal{G}_{\varphi}|$ represents the average number of goals in the datasets, and $|Obs|$ the average number of observations. Each row in these inner tables represents the observation level, varying from 10\% to 100\%. We note that we are unable to provide a \textit{direct comparison} of our approach against existing recognition approaches in the literature because most of these approaches perform a non-trivial process that transforms a recognition problem into planning problems to be solved by a planner~\cite{RamirezG_AAAI2010,Sohrabi_IJCAI2016}. Even adapting such a transformation to work in \FOND settings with temporally extended goals, we cannot guarantee that it will work properly in the problem setting we propose in this paper. 

\subsubsection{Conjunctive and Eventuality Goals.}

The first inner table shows the average results comparing the performance of our approach between conjunctive goals and temporally extended goals using the temporal operator $\Diamond$ (\textit{eventually}). We refer to this comparison as \textit{baseline}, since these two types of goals have the same semantics. We can see that the results for these two types of goals are very similar for all metrics. Moreover, it is also possible to see that our recognition approach is very accurate and performs well at all levels of observability, yielding high \textit{TPR} values, and low \textit{FPR} and \textit{FNR} values for more than 10\% of observability. Note that for 10\% of observability and \LTLf goals for $\Diamond\varphi$, the \textit{TPR} average value is 0.74, and it means for 74\% of the recognition problems our approach recognized correctly the intended temporally extended goal when observing, on average, only 3.85 actions. 

\subsubsection{Results for \LTLf Goals.} 

Regarding the results for the two types of \LTLf goals (second inner table), we can see that our approach shows to be accurate for all metrics at all levels of observability, apart from the results for 10\% of observability for \LTLf goals in which the formulas must be recognized in a certain order. Note that even when observing just a few actions (2.1 for 10\% and 5.4 for 30\%) our approach is accurate, but not as accurate as for more than 30\% of observability.

\subsubsection{Results for \PLTLf Goals.}

Finally, as for the results for the two types of \PLTLf goals, we can see in the last inner table that, the overall average number of observations $|Obs|$ is less than the average for the other datasets, making the task of goal recognition more difficult for the \PLTLf datasets. Yet, we can see that our recognition approach remains accurate when dealing with fewer observations. We can also see that the values of \textit{FNR} increase for low observavility, but the \textit{FPR} values, are, on average, inferior to $\approx$ 0.15.

\begin{table}[!ht]
\centering
\fontsize{7.1}{6.1}\selectfont
\setlength\tabcolsep{1.5pt}
\begin{tabular}{ccccccccccccl}
\toprule	
    &                      &         &  & \multicolumn{4}{c}{\thead{\scriptsize Conjunctive Goals \\ $\phi_1\land \phi_2$}} &  & \multicolumn{4}{c}{\thead{\scriptsize \LTLf Eventuality Goals \\ $\Diamond\phi$}} \\ \hline
    & $|\mathcal{G}_{\phi}|$            & $|Obs|$ &  & \emph{Time}    & \emph{TPR}     & \emph{FPR}     & \emph{FNR}     &  & \emph{Time}      & \emph{TPR}      & \emph{FPR}      & \emph{FNR}      \\ \hline
10  & \multirow{5}{*}{5.2} & 3.85     &  & 189.1     & 0.75    & 0.15    & 0.25    &  & 243.8       & 0.74     & 0.11     & 0.26     \\
30  &                      & 10.7     &  & 187.2     & 0.85    & 0.08    & 0.15    &  & 235.1       & 0.86     & 0.10     & 0.14     \\
50  &                      & 17.4     &  & 188.4     & 0.83    & 0.09    & 0.17    &  & 242.1       & 0.89     & 0.07     & 0.11     \\
70  &                      & 24.3     &  & 187.8     & 0.86    & 0.08    & 0.14    &  & 232.1       & 0.92     & 0.08     & 0.08     \\
100 &                      & 34.7     &  & 190.4     & 0.85    & 0.09    & 0.15    &  & 272.8       & 0.95     & 0.09     & 0.05   \\
\end{tabular}

\begin{tabular}{ccccccccccccl}
\toprule	
    &                      &         &  & \multicolumn{4}{c}{\thead{\scriptsize \LTLf Ordering Formulas \\ $\Diamond(\phi_1 \land \Next(\Diamond\phi_2))$}} &  & \multicolumn{4}{c}{\thead{\scriptsize \LTLf Goals using Until \\ $\phi_1 \Until \phi_2$}} \\ \hline
    & $|\mathcal{G}_{\phi}|$            & $|Obs|$ &  & \emph{Time}    & \emph{TPR}     & \emph{FPR}     & \emph{FNR}     &  & \emph{Time}      & \emph{TPR}      & \emph{FPR}      & \emph{FNR}      \\ \hline
10  & \multirow{5}{*}{4.0} & 2.1     &  & 136.1     & 0.68    & 0.15    & 0.32    &  & 217.9       & 0.79     & 0.11     & 0.21     \\
30  &                      & 5.4     &  & 130.9     & 0.84    & 0.13    & 0.16    &  & 215.8       & 0.91     & 0.12     & 0.09     \\
50  &                      & 8.8     &  & 132.1     & 0.88    & 0.10    & 0.12    &  & 210.1       & 0.93     & 0.10     & 0.07     \\
70  &                      & 12.5    &  & 129.2     & 0.95    & 0.06    & 0.05    &  & 211.5       & 0.97     & 0.09     & 0.03     \\
100 &                      & 17.1    &  & 126.6     & 0.94    & 0.05    & 0.06    &  & 207.7       & 0.97     & 0.07     & 0.03     \\
\end{tabular}

\begin{tabular}{ccccccccccccl}
\toprule	
    &                      &         &  &
    \multicolumn{4}{c}{\thead{\scriptsize \PLTLf Goals using Once  \\ $\phi_1 \land \past \phi_2$}} &  & \multicolumn{4}{c}{\thead{\scriptsize \PLTLf Goals using Since \\ $\phi_1 \land (\lnot\phi_2 \Since \phi_3)$}} \\ \hline
    & $|\mathcal{G}_{\phi}|$            & $|Obs|$ &  & \emph{Time}    & \emph{TPR}     & \emph{FPR}     & \emph{FNR}     &  & \emph{Time}      & \emph{TPR}      & \emph{FPR}      & \emph{FNR}      \\ \hline
10  & \multirow{5}{*}{4.0} & 1.7     &  & 144.8     & 0.73    & 0.11    & 0.27    &  & 173.5     & 0.76    & 0.18    & 0.24     \\
30  &                      & 4.6     &  & 141.3     & 0.84    & 0.07    & 0.16    &  & 173.3     & 0.87    & 0.12    & 0.13     \\
50  &                      & 7.3     &  & 141.9     & 0.89    & 0.08    & 0.11    &  & 172.9     & 0.85    & 0.09    & 0.15     \\
70  &                      & 10.3    &  & 142.9     & 0.95    & 0.07    & 0.05    &  & 171.1     & 0.97    & 0.07    & 0.03     \\
100 &                      & 14.2    &  & 155.8     & 0.97    & 0.07    & 0.03    &  & 169.3     & 0.94    & 0.02    & 0.06   \\
\bottomrule
\end{tabular}

\caption{Results for Conjunctive, \LTLf, and \PLTLf goals.}
\label{tab:gr_results_conjunctive_event}
\end{table}


\section{Conclusions}

In this paper, we have studied goal recognition for temporally extended goals, specified in both \LTLf and \PLTLf, in \FOND planning domain models. We also have developed a novel probabilistic recognition framework in such a setting. 
On a practical perspective, we have implemented a compilation of temporally extended goals that allows us to reduce the problem of \FOND planning for \LTLf/\PLTLf goals to standard \FOND planning, and have performed extensive experiments that show the feasibility of the approach.
Indeed, we have shown that, for all evaluated domains, our recognition approach yields high accuracy for recognizing temporally extended goals at several levels of observability, for both \LTLf and \PLTLf goals. As future work, we intend to address noisy observations, 
and recognize not only the temporal goal that the observed agent aims to achieve, but also anticipate the sequence of actions that the agent is executing.


\section*{Acknowledgments}
Research partially supported by the ERC Advanced Grant WhiteMech (No. 834228) and by the EU ICT-48 2020 project TAILOR (No. 952215).

\bibliography{journal-abbreviations,references}

\end{document}